\documentclass[sigconf]{acmart}

\graphicspath{{image/}}

\AtBeginDocument{%
  }

\setcopyright{acmlicensed}
\copyrightyear{2027}
\acmYear{2027}
\acmDOI{XXXXXXX.XXXXXXX}
\acmConference[KDD '27]{Proceedings of the 33rd ACM SIGKDD Conference on Knowledge Discovery and Data Mining}{August 2027}{San Jose, CA, USA}
\acmBooktitle{Proceedings of the 33rd ACM SIGKDD Conference on Knowledge Discovery and Data Mining (KDD '27), August 2027, San Jose, CA, USA}
\acmISBN{978-1-4503-XXXX-X/2027/08}

\usepackage{booktabs}   %
\usepackage{multirow}
\usepackage{subcaption}
\usepackage{xspace}
\usepackage{enumitem}   %

\usepackage{tikz}
\usetikzlibrary{positioning,arrows.meta,shapes.geometric,calc,fit,backgrounds,
  decorations.pathreplacing}

\definecolor{seamblue}{RGB}{46,116,181}    %
\definecolor{seamblubg}{RGB}{224,236,247}  %
\definecolor{seamgraybg}{RGB}{238,238,240} %
\definecolor{seamgreenbg}{RGB}{224,240,228}%
\definecolor{edgetmp}{RGB}{31,119,180}     %
\definecolor{edgecha}{RGB}{214,123,32}     %
\definecolor{edgescn}{RGB}{44,140,84}      %

\tikzset{
  seamfont/.style={font=\footnotesize},
  flowbox/.style={rectangle, rounded corners=2pt, draw=black!55,
    fill=seamgraybg, align=center, inner sep=4pt, minimum height=8mm,
    font=\footnotesize},
  seambox/.style={flowbox, draw=seamblue, fill=seamblubg, very thick},
  iobox/.style={rectangle, rounded corners=2pt, draw=black!45,
    fill=seamgreenbg, align=center, inner sep=3pt, font=\scriptsize},
  statenode/.style={circle, draw=black!55, fill=seamgraybg,
    minimum size=7mm, font=\scriptsize, inner sep=0pt},
  seedstate/.style={statenode, draw=seamblue, fill=seamblubg, very thick},
  flowarr/.style={-{Latex[length=2mm]}, draw=black!65, thick},
  etmp/.style={-{Latex[length=1.6mm]}, draw=edgetmp, thick},
  echa/.style={-{Latex[length=1.6mm]}, draw=edgecha, thick, dashed},
  escn/.style={-{Latex[length=1.6mm]}, draw=edgescn, thick, dotted},
}

\makeatletter
\@ifundefined{footinsauthorsaddresses}{}{%
  \setlength{\skip\footinsauthorsaddresses}{4pt}}   %
\@ifundefined{footinscopyrightpermission}{}{%
  \setlength{\skip\footinscopyrightpermission}{4pt}}%
\def\footnoterule{\kern-3\p@\hrule\@width 4pc\kern 2\p@}
\makeatother

\newcommand{\sysname}{SEAM\xspace}              %
\newcommand{\pipename}{SEAM-Agent\xspace}        %
\newcommand{\benchname}{SEAM-Bench\xspace}       %

\newcommand{\tabscalebench}{1.0}     %
\newcommand{\tabscalehone}{1.0}      %
\newcommand{\tabscaletargeted}{1.0}  %
\newcommand{\tabscaleblind}{1.0}      %
\newcommand{\tabscalehtwo}{0.92}     %
\newcommand{\tabscaleoffline}{1.0}   %
\newcommand{\tabscaleabbrev}{1.0}    %
\newcommand{\tabscalestatedims}{1.0} %
\newcommand{\tabscaleadoption}{1.0}     %

\begin{document}

\title{\sysname: Shot Entity-Attribute Memory for Consistent
  Short-Drama Generation at Scale}

\author{Jiaqi Liu}
\affiliation{%
  \institution{Shanghai Jiao Tong University}
  \city{Shanghai}
  \country{China}
}
\email{jkliu189@gmail.com}

\author{Maolin Ran}
\affiliation{%
  \institution{Shanghai Jiao Tong University}
  \city{Shanghai}
  \country{China}
}
\email{maolinr03@sjtu.edu.cn}

\author{Xiaoyang Lu}
\affiliation{%
  \institution{Shanghai Jiao Tong University}
  \city{Shanghai}
  \country{China}
}
\email{xiaoyangl@sjtu.edu.cn}

\author{Jian Wang}
\affiliation{%
  \institution{CreativeFitting}
  \city{Shanghai}
  \country{China}
}
\email{jim.wang@creativefitting.ai}

\author{Weiwen Liu}
\affiliation{%
  \institution{Shanghai Jiao Tong University}
  \city{Shanghai}
  \country{China}
}
\email{wwliu@sjtu.edu.cn}
\authornote{Corresponding author.}

\author{Jianghao Lin}
\affiliation{%
  \institution{Shanghai Jiao Tong University}
  \city{Shanghai}
  \country{China}
}
\email{linjianghao@sjtu.edu.cn}

\author{Yong Yu}
\affiliation{%
  \institution{Shanghai Jiao Tong University}
  \city{Shanghai}
  \country{China}
}
\email{yyu@sjtu.edu.cn}

\author{Weinan Zhang}
\affiliation{%
  \institution{Shanghai Jiao Tong University}
  \city{Shanghai}
  \country{China}
}
\email{wnzhang@sjtu.edu.cn}
\authornotemark[1]

\renewcommand{\shortauthors}{Liu et al.}

\begin{abstract}
Short-drama generation has grown into a large, industrialized pipeline, and as it
scales from isolated shots to the episode level, visual continuity has become a
critical bottleneck. Current agent frameworks generate each shot in isolation, so
context drifts across shots, and props, character posture, and blocking turn
inconsistent. Once assembled, these small discrepancies amplify into severe visual
breaks. We present \sysname (\textbf{S}hot
\textbf{E}ntity-\textbf{A}ttribute \textbf{M}emory), a training-free,
model-agnostic memory graph that repairs continuity entirely at the prompt-text
layer by extracting a multi-dimensional state for every shot, retrieving only
causally prior context over the resulting graph, filtering it selectively, and
injecting the surviving constraints by natural-language prompt rewriting. We further release
\benchname, a double-blind continuity storyboarding benchmark, on which \sysname
raises cross-episode continuity recall from $0.700$ to $0.946$, generalizes
across six mainstream text models, and yields consistent, though not yet
significant, gains at the generated-image layer. Deployed as a mandatory stage in
CreativeFitting's \pipename production pipeline over 201 shots, \sysname reaches a
$96.5\%$ director-acceptance rate with zero unsafe injections; a conservative
counterfactual attributes at least $21.9$ percentage points of that rate to its
cross-episode memory.
\end{abstract}

\ccsdesc[500]{Computing methodologies~Natural language generation}
\ccsdesc[300]{Computing methodologies~Image and video generation}
\ccsdesc[300]{Information systems~Retrieval models and ranking}

\keywords{Memory graph, Short-drama generation,
  Retrieval augmentation}

\maketitle

\section{Introduction}
\label{sec:intro}

Short drama, a form of vertical, fast, multi-episode video fiction, is one of the
fastest-growing segments of digital entertainment, and its production is now
being reshaped by generative
AI~\cite{movieagent2025,kubrick2024,cineagi2026}. Its industrial workflow is
highly standardized, spanning script writing, storyboarding, keyframe
generation, and video synthesis, and agent frameworks increasingly absorb the
repetitive labor around a professional director rather than replace the creative
core~\cite{directorllm2024}. The commercial stakes are already
concrete: fully AI-generated titles now compete with live-action ones for the same
audience, as on CreativeFitting's Reel.AI, one such platform serving AI-generated
short dramas worldwide and the production setting we study in this paper. What
makes the domain distinctive is
its scale: a single title routinely spans tens of episodes with tens of shots
each, so one production comprises thousands of shots that must remain mutually
coherent.

\begin{figure}[t]
  \centering
  \includegraphics[width=\columnwidth]{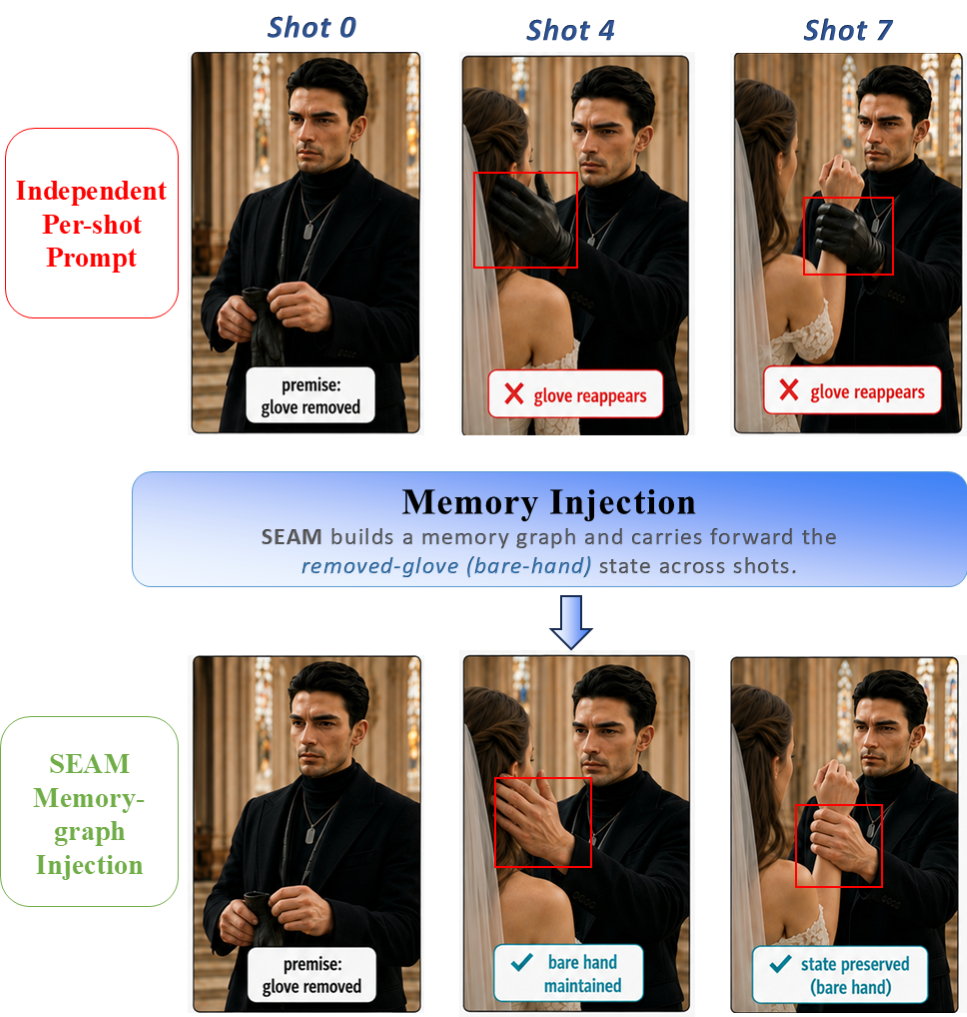}
  \Description{Two rows of three keyframes for shots 0, 4, and 7. In the top
    row, generated per shot independently, a glove removed in shot 0 reappears
    in shots 4 and 7. In the bottom row, generated with memory injection, the
    bare-hand state is preserved in both later shots.}
  \caption{Cross-shot continuity break vs.\ \sysname repair. Top: per-shot
    independent generation loses an earlier state (a removed glove reappears),
    a contradiction no single shot can catch. Bottom: \sysname injects the
    prior state from its memory graph into the shot prompt, carrying it forward
    so the contradiction disappears.}
  \label{fig:teaser}
\end{figure}

At this scale, \emph{visual continuity} is the central bottleneck, because each
shot's prompt is generated in isolation and carries no persistent state memory.
As shown in Figure~\ref{fig:teaser}, per-shot keyframe generation loses the
glove that Shot~0 removed and lets it reappear in Shots~4 and~7; injecting the
prior state from \sysname's memory graph carries the bare-hand state forward, and
both shots stay consistent. Generators follow their prompts faithfully, so they
amplify rather than absorb such contradictions, and over thousands of shots the
drift accumulates into visual breaks a human editor must repair by hand.

Prior work enforces consistency inside a specific
generator~\cite{holocine2025,memflow2025,lightscamera2025,recap2026}, anchors it
to 3D scenes or to video generated end to
end~\cite{storyblender2026,stage2026}, or reuses generic retrieval and memory
systems built for factual question
answering~\cite{ketrag2025,assomem2025,telemem2025}. All of them leave the
storyboard prompt layer unaddressed, yet that layer is where the repair belongs:
it sits upstream of every generator and stays readable and editable by the
director. Repairing continuity there, inside a live production pipeline, raises
three challenges that existing methods do not resolve.
\emph{C1: Heterogeneous, evolving visual state.} A shot's continuity depends on
many dimensions at once (scene, atmosphere, characters and their states, spatial
blocking, props, actions, camera style), each carrying forward or changing
independently as the story progresses; unstructured text memory cannot represent
this structured, multi-dimensional state.
\emph{C2: Causal and selective reuse.} Only \emph{prior} state may be reused, so
that causality is preserved, and only the truly relevant fragments should be
injected, since indiscriminate copying overrides the director's creative intent.
\emph{C3: Generality and pluggability.} Generative backbones iterate rapidly, so
a deployable remedy cannot be rebuilt whenever the underlying image or video
model is replaced. It must be training-free, model-agnostic, pluggable as a
standalone stage, and able to add continuity without altering the shot grammar
or visual intent the director specified.

Guided by these challenges, we cast continuity as a
\emph{retrieval-and-injection} problem over cross-shot shared state. For
\emph{C1}, we parse each shot into a multi-dimensional state node and connect
nodes by temporal-adjacency, character-co-occurrence, and scene-co-occurrence
edges, turning the evolving state into an explicit, queryable memory graph. For
\emph{C2}, we retrieve only the relevant prior state and let an LLM selectively
decide which candidates to inject; the visual description is then rewritten
naturally, so continuity is added while the shot grammar and the director's
intent stay intact. For \emph{C3}, the whole procedure operates purely at the
prompt-text layer and touches neither the model weights nor the downstream
backbone, so it plugs in as a standalone stage and keeps working unchanged when
the image or video model is swapped out. We call this framework \sysname, a
\textbf{S}hot \textbf{E}ntity-\textbf{A}ttribute \textbf{M}emory graph, and embed
it as one stage of \pipename, a multi-agent storyboarding pipeline that any
script-to-screen system can adopt and that we run in commercial production.

\noindent\textbf{Contributions.} This paper makes the following contributions:
\begin{itemize}[leftmargin=10pt, topsep=2pt, itemsep=2pt, parsep=0pt]
  \item \emph{We introduce \sysname, the first shot entity-attribute memory
        graph for storyboard-layer continuity.} To our knowledge this is the
        first work to repair short-drama continuity at the storyboard prompt
        layer: each shot becomes a multi-dimensional visual state node, the
        temporal, character, and scene relations governing continuity become
        typed edges, and cross-shot consistency becomes a
        retrieval-and-injection problem over this graph.
  \item \emph{We instantiate it as a training-free, prompt-text-layer
        procedure} of three stages (state extraction with graph construction,
        causal retrieval with selective filtering, natural-rewrite injection)
        that touches neither model weights nor the downstream backbone, so the
        stage stays model-agnostic and survives backbone upgrades.
  \item \emph{We release \benchname, a double-blind continuity storyboarding
        benchmark}\footnote{\url{https://huggingface.co/datasets/Jackyqq/SEAM-Bench}}
        that standardizes continuity evaluation for industrial short-drama
        generation at both the prompt and image layers.
  \item \emph{We validate \sysname in live commercial production.} Deployed as a
        mandatory stage of the \pipename pipeline in CreativeFitting's
        production system, it reaches a $96.5\%$ director-acceptance rate over 201
        shots, cutting the manual continuity inspection a polished AI short drama
        otherwise demands from directors---evidence that the design holds outside
        the benchmark.
\end{itemize}

Section~\ref{sec:related} surveys related work;
Section~\ref{sec:prelim} formalizes the problem and the continuity criterion;
Section~\ref{sec:method} presents \sysname, its memory graph, and its integration
into \pipename;
Section~\ref{sec:exp} reports \benchname, the three-layer evaluation, and the
online deployment evidence.

\section{Related Work}
\label{sec:related}

\subsection{Agentic Short-Drama Production}
\label{sec:rel_pipeline}
The rise of short drama has motivated dedicated datasets, generation pipelines,
and evaluation protocols. Large script corpora pair screenplays with shooting
scripts~\cite{skyscript2024}, while agents take complementary creative roles:
director--actor collaboration for controllable script writing~\cite{ibsen2024},
personalized frameworks carrying a single sentence to a produced
drama~\cite{onesentence2026}, and geometry-guided control of
cinematography~\cite{dramadirector2026}. The same role decomposition scales to
feature-length work, where multi-agent movie pipelines distribute director,
screenwriter, and storyboard-artist roles across LLM or VLM agents to plan and
render multi-scene video~\cite{movieagent2025,kubrick2024,cineagi2026}, with
parallel efforts supplying synchronized sound~\cite{reelwave2025}. How to score
the result is itself unsettled: these systems predominantly report human Likert
ratings or a single LLM judge~\cite{movieagent2025,directorllm2024}, leaving
conclusions resting on one evaluator family, whereas story-visualization benchmarks
supply image-side protocols for character identity and scene
consistency~\cite{vistorybench2025,consistory2024,vbench2024} and
storyboard-level benchmarks have begun to probe cinematographic
quality~\cite{lstoryboard2025,cmlbench2025}.
\textit{Delta.} These pipelines automate authoring and rendering end to end but
treat consistency as a byproduct of the generator or of scene-level planning,
leaving no explicit representation of what carries over between shots, and no
benchmark evaluates \emph{cross-episode} continuity at the storyboard text layer
alongside the keyframes rendered from it. We isolate and repair continuity at
the textual storyboard that directors and downstream tools actually consume, via
a shot-level memory graph deployed as a pipeline stage rather than a standalone
generator, and release \benchname to check a continuity claim at the prompt
layer and the image layer at once.

\subsection{Memory for Cross-Shot Consistency}
\label{sec:rel_consistency}
A large body of work seeks visual consistency across shots or frames. One line
bakes it into the \emph{generator}: cinematic models render coherent multi-shot
sequences in a single pass~\cite{holocine2025}, memory-flow methods propagate
state across long-video generation~\cite{memflow2025}, and diffusion extensions
carry identity through transition tokens, caches, or entity-grounded
scheduling~\cite{shotadapter2025,filmweaver2025,groundshot2026}.
A second anchors generation to an external structure such as a storyboard or a
memory pack for 3D scenes~\cite{storyblender2026,stage2026}. A third works in the
image domain, preserving character identity across story panels by extracting a
reusable character~\cite{chosenone2023}, controlling multi-character
layout~\cite{talecrafter2023}, interleaving text with images~\cite{seedstory2024},
grounding identity referentially~\cite{recap2026}, or repairing inconsistent
panels after the fact~\cite{auditrepair2025}; that such anchoring is load-bearing
is confirmed by character-stable pipelines reporting catastrophic drops once it is
removed~\cite{lightscamera2025}. A separate line asks instead \emph{how} an
agent should store and retrieve what it has seen, arguing that episodic
memory is the missing piece over long
horizons~\cite{memorysurvey2024,episodic2025} and structuring retrieval through
graph-based RAG~\cite{ketrag2025}, associative reuse~\cite{assomem2025}, or
persistent multimodal memory~\cite{telemem2025,memoria2025}, with continuity
checked reference-free through entity graphs~\cite{entitygraph2013} or NLI
contradiction detection~\cite{summac2022}.
\textit{Delta.} The first three lines operate on pixels, latents, or a specific
generator and are largely training-based or backbone-coupled, hard to insert
into a model-heterogeneous pipeline; the memory machinery is instead built for
factual recall. Visual continuity differs in what must be stored and in what may
be reused: the unit of memory is a shot's multi-dimensional visual state rather
than a proposition, the relations governing reuse are temporal adjacency and
character/scene co-occurrence rather than semantic relatedness, and reuse must be
strictly causal. \sysname adopts the retrieval-and-injection view but instantiates
it over these shot-level states one level earlier than the generator, at the
storyboard prompt-text layer, staying training-free and model-agnostic.

\section{Preliminaries}
\label{sec:prelim}

This section fixes the notation for short dramas and shots, states the
generation task, and defines the continuity criterion our core metric is built
on. The memory graph is the heart of our design rather than background, so it is
defined where it is used (Section~\ref{sec:method_extract}).

\paragraph{Dramas and shots.}
A short drama is an episode-ordered set $\mathcal{D}=\{E_1,\dots,E_M\}$ of $M$
episodes, whose $m$-th episode $E_m=(s_1,\dots,s_{n_m})$ is a temporally ordered
sequence of $n_m$ shots ($1\le m\le M$). Each shot $s_i$ is a structured tuple
\begin{equation}
  s_i=(\ell_i,\ \kappa_i,\ x_i,\ t_i,\ C_i,\ P_i),
  \label{eq:shot}
\end{equation}
where $\ell_i\in\mathcal{L}$ is the scene/location, $\kappa_i$ the shot grammar
(shot size, angle, movement), $x_i$ the visual description, $t_i$ the dialogue,
and $C_i\subseteq\mathcal{C}$, $P_i\subseteq\mathcal{P}$ the characters and props
active in the shot. Here $\mathcal{C}$, $\mathcal{P}$ and $\mathcal{L}$ are the
drama-wide universes of characters, props and scenes, respectively; the shot
index $i$ runs within an episode, and where cross-episode context matters it is
read in the global shot order that $\mathcal{D}$ induces.

\paragraph{Generation task.}
The pipeline produces, for each shot, a prompt $p_i$ that a downstream
image/video generator consumes. Writing $f_\theta$ for the shotlist-authoring
model with parameters $\theta$ (an LLM backbone in all our experiments), the
per-shot, stateless baseline is
\begin{equation}
  p_i=f_\theta(s_i),
  \label{eq:baseline}
\end{equation}
that is, the prompt depends only on the current shot and carries no cross-shot
state. This is the condition our memory stage is measured against.

\paragraph{Continuity and continuity recall.}
Continuity concerns the \emph{persistent entities} of a drama---its characters,
props and scenes---whose visual attributes should stay consistent across shots
unless the narrative motivates a change. For an entity
$e\in\mathcal{C}\cup\mathcal{P}\cup\mathcal{L}$ we write $\pi_i(e)$ for its
\emph{state projection}: the visual attributes (character appearance, prop
possession, scene layout) that shot $s_i$ ascribes to $e$, with
$\pi_i(e)=\varnothing$ when $e$ is inactive in that shot. For a shot pair $i<j$
in which $e$ is active in both, a \emph{continuity defect} $\delta(e,i,j)=1$ is
recorded when $\pi_i(e)$ contradicts $\pi_j(e)$ without narrative motivation.
This grounds the \emph{continuity recall}
\begin{equation}
  \mathrm{Recall}=
    \frac{\#\{\text{defects successfully repaired}\}}
         {\#\{\text{defects judged in need of repair}\}},
  \label{eq:recall}
\end{equation}
our core metric for repairing cross-shot and cross-episode continuity
(measured in Section~\ref{sec:exp}).

\section{Methodology}
\label{sec:method}

In this section, we introduce our \textbf{S}hot \textbf{E}ntity-\textbf{A}ttribute \textbf{M}emory graph for short-drama
storyboarding (i.e., \sysname).

\subsection{Overview}
\label{sec:method_overview}

Guided by the three challenges of Section~\ref{sec:intro} (\emph{C1}
heterogeneous evolving state, \emph{C2} causal selective reuse, \emph{C3}
training-free model-agnostic delivery), our methodology has two parts:
\emph{how continuity is represented and repaired}, and \emph{how the repair is
delivered inside a live production system}. For the first, we design \sysname, a
cross-shot continuity memory graph that turns the retrieval-and-injection view of
continuity into an executable, prompt-text-layer procedure: from the episode
\emph{script} we extract a structured multi-dimensional visual state for every
shot and link the states into a directed memory graph whose temporal, character,
and scene edges govern continuity; when the shotlist is (re-)authored, we retrieve
only the causally prior context, filter it selectively to discard redundant or
conflicting fragments, and rewrite the shot's visual description so the surviving
continuity is woven in while the shot grammar and the director's intent stay
intact. \sysname runs in an \emph{online} LLM-driven mode, used for every result
reported in this paper, and a deterministic \emph{offline} mode that substitutes
rules for each LLM call so the stage still runs where no model is reachable. For the
second, we embed \sysname into \pipename, our multi-agent storyboarding pipeline,
as a mandatory ``memory-optimizer'' stage that builds the graph from the script,
consumes the upstream shotlist, and repairs continuity shot by shot before
rendering; because it reads the script and rewrites only the textual shotlist, it
depends on neither the upstream authoring model nor the downstream backbone.

\begin{figure*}[t]
  \centering
  \includegraphics[width=\textwidth]{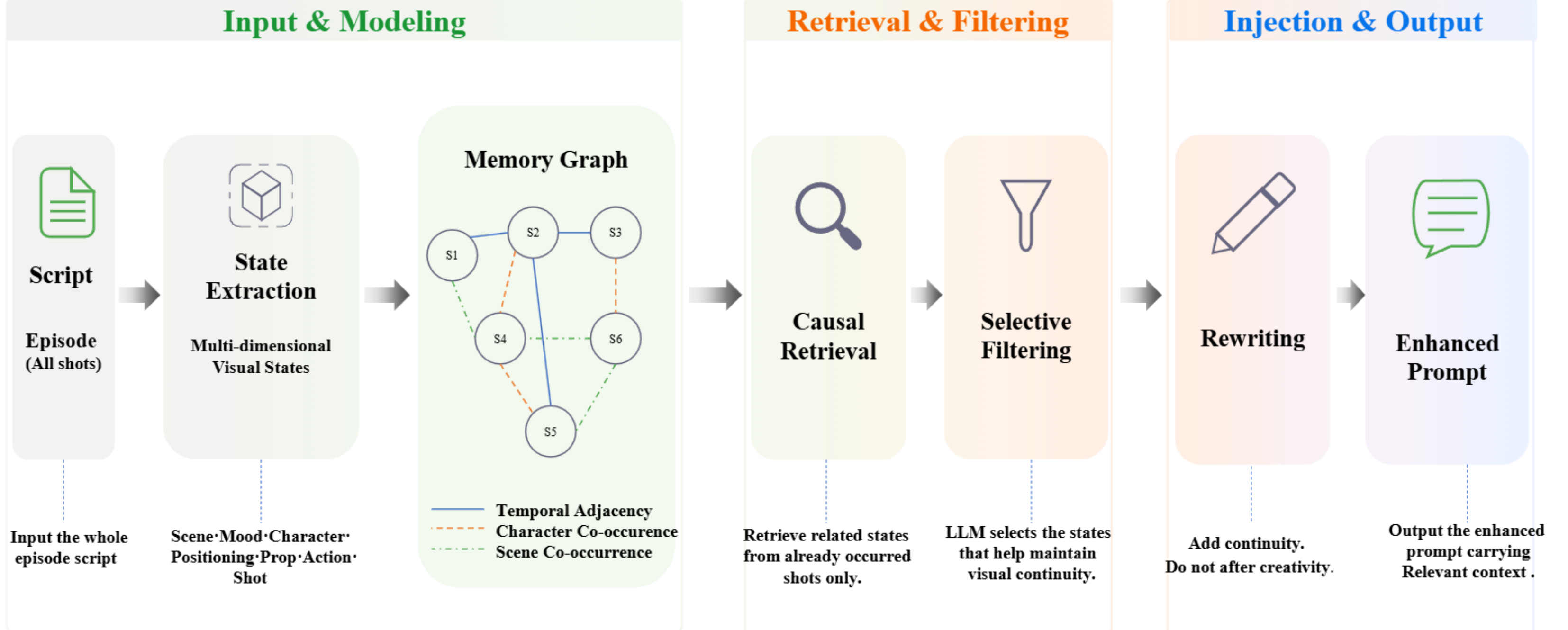}
  \Description{Block diagram of the three SEAM stages: the script is parsed into
    multi-dimensional shot states, the states are linked into a memory graph,
    and prior context is retrieved, filtered, and rewritten into each shot's
    visual description.}
  \caption{\sysname overview. The episode script is parsed into
    multi-dimensional shot states, which are linked into a memory graph by
    temporal, character, and scene edges. For each shot of the upstream shotlist,
    causally prior context is retrieved, filtered selectively, and rewritten into
    the visual description, yielding a continuity-enriched prompt.}
  \label{fig:overview}
\end{figure*}

\subsection{\sysname: A Cross-Shot Continuity Memory Graph}
\label{sec:method_framework}

As shown in Figure~\ref{fig:overview}, \sysname is organized as three stages
operating entirely at the prompt-text layer: state extraction and graph
construction, graph retrieval with selective filtering, and natural-rewrite
injection. We describe each in turn.

\subsubsection{State Extraction and Graph Construction}
\label{sec:method_extract}
Whereas the state projection $\pi_i$ of Section~\ref{sec:prelim} tracks one
entity at a time, our memory represents a whole shot at once. The extractor takes
the episode script as input and, for every shot $s_i$ it induces
(Eq.~\eqref{eq:shot}), produces a $d$-dimensional visual state
$\boldsymbol{\sigma}_i=(\sigma_i^{1},\dots,\sigma_i^{d})$, whose $d=8$ dimensions
cover scene, atmosphere, characters, character states, spatial blocking, props,
actions, and camera style (Table~\ref{tab:state_dims}). We formulate the
extraction as an operator
\begin{equation}
  \boldsymbol{\sigma}_i=\Phi(s_i),
  \label{eq:extract}
\end{equation}
which is instantiated in the online mode by prompting an LLM with a fixed
dimension template that returns one structured record per shot, and, in the
offline mode, by deterministic rule-based parsing of the same script.
Heterogeneous sources are first normalized into a unified shot stream, so $\Phi$
applies uniformly regardless of the source format, and the dimension set is
configurable rather than fixed by the formulation; $d=8$ is the instantiation
used throughout this paper.

\begin{figure}[t]
  \centering
  \includegraphics[width=\columnwidth]{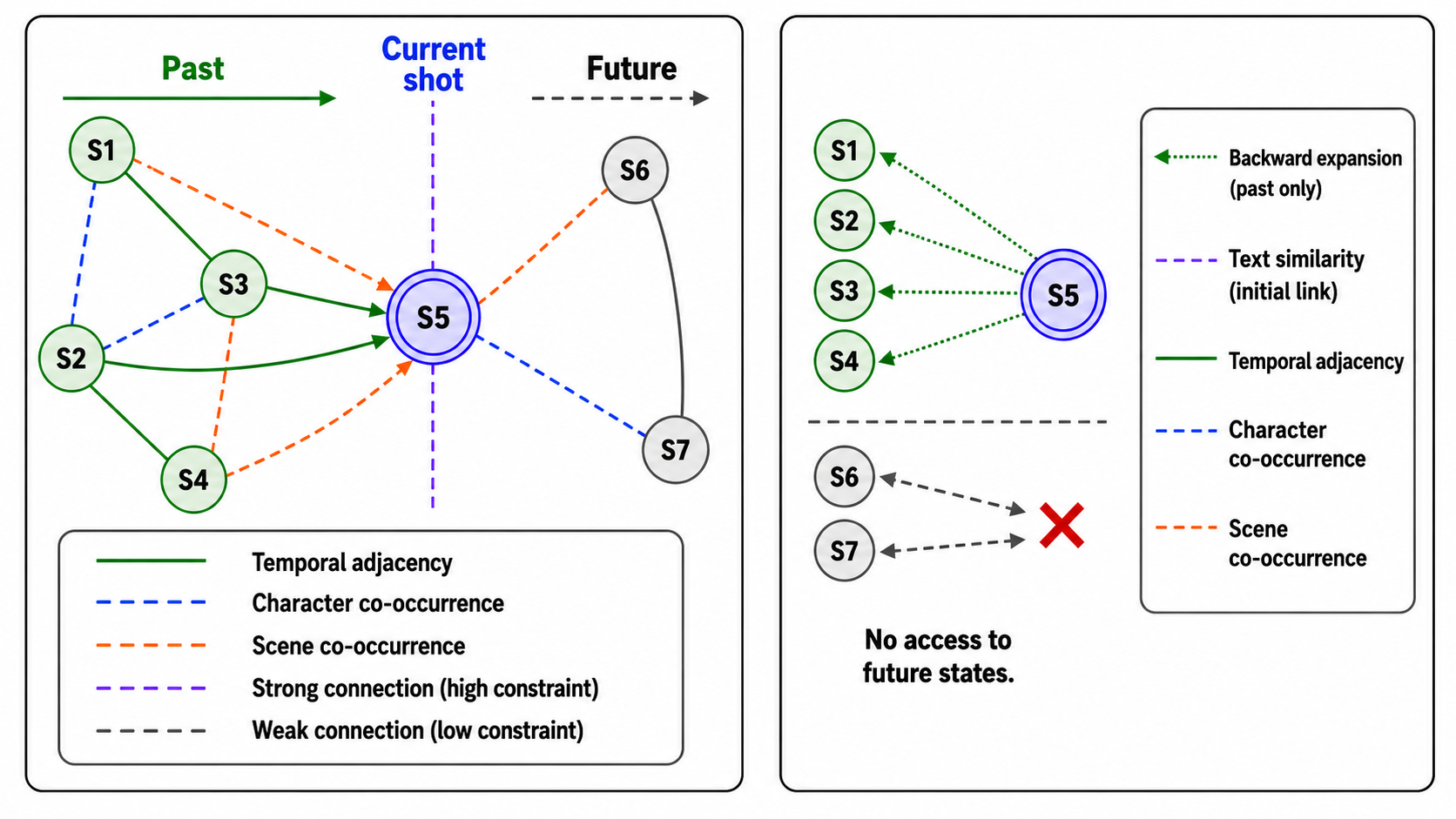}
  \Description{Two panels: the left shows shot nodes joined by temporal,
    character, and scene edges; the right shows retrieval expanding backward
    from seed nodes over the earlier part of the graph only.}
  \caption{(a) The three relation types (temporal, character, and scene). (b)
    Graph retrieval with backward edge expansion over the prior subgraph
    $G_{<i}$.}
  \label{fig:edges}
\end{figure}

Over the resulting states we construct a directed \emph{memory graph} $G=(V,E)$.
Its nodes $V$ are the shot-state nodes $v_i$ (each carrying
$\boldsymbol{\sigma}_i$) together with resource-entity nodes materialized for the
recurring characters, props, and scenes, and we
connect nodes through the three relation types that carry continuity: a
temporal-adjacency relation linking a shot to its neighbors within a window
$|i-j|\le N$, a character-co-occurrence relation linking shots that share a
character ($C_i\cap C_j\neq\varnothing$), and a scene-co-occurrence relation
linking shots set in the same place ($\ell_i=\ell_j$). Each edge $(i,j)$ carries a
continuity weight $w_{ij}\in[0,1]$ that grades how strongly the earlier shot
constrains the later one, from a hard must-continue link ($w_{ij}=1$) down to a
weak association. In the online mode the relation type and weight of every edge
are inferred by an LLM over overlapping shot windows and then deduplicated by
keeping, for each ordered pair and type, the edge of maximum weight; in the
offline mode the same three relations are recovered by deterministic
co-occurrence and adjacency rules. These relations jointly determine which
earlier shots are eligible to constrain the current one
(Figure~\ref{fig:edges}(a)), and the weights $w_{ij}$ drive the weighted,
multi-hop expansion used during retrieval.

\subsubsection{Graph Retrieval and Selective Filtering}
\label{sec:method_retrieve}
To repair shot $s_i$, we retrieve a continuity context
$\mathcal{K}_i=\mathrm{Retrieve}(G_{<i},s_i)$ from the \emph{prior} subgraph
$G_{<i}$, the part of $G$ induced by nodes with index $j<i$. This enforces
causality: a shot may inherit state from what has already happened, never from
the future, so no forward leakage can occur.

We support two complementary retrieval modes. When a shot carries explicit
resource references, we take its prior state by \emph{resource overlap}: writing
$R_i=C_i\cup P_i\cup\{\ell_i\}$ for its resource set, we collect the prior nodes
$\{v_j\in G_{<i}\mid R_j\cap R_i\neq\varnothing\}$ and keep the most recent
states of the shared characters, props, and scenes; when no prior node overlaps,
we fall back to the last few preceding nodes so the context is never empty.
For arbitrary inputs without such references, we instead perform
\emph{text-matched retrieval with edge expansion}, first scoring every prior node
$v_j\in G_{<i}$ by a lightweight lexical overlap with $s_i$,
\begin{equation}
  \mathrm{sim}(s_i,v_j)=\min\!\Big(1,\;
    \frac{|T(s_i)\cap T(v_j)|}{|T(s_i)|}
    +\beta[\,C_i\cap C_j\neq\varnothing\,]\Big),
  \label{eq:sim}
\end{equation}
where $T(\cdot)$ is a token set that mixes whole words with character-level
bigrams so that the score stays robust across languages, the first term is the recall of the
query tokens, and the second term adds a bonus $\beta$ when the two shots share a
named character. We keep the top-$k$ prior nodes with $\mathrm{sim}>0$ as seeds
$\mathcal{S}_i$, and then expand $h$ hops \emph{backward} along the temporal,
character, and scene relations, traversing only edges whose weight clears a
threshold $w_{ij}\ge\tau$,
\begin{equation}
  \mathcal{K}_i=\bigcup_{v\in\mathcal{S}_i}
    \mathrm{Expand}_{\le h}^{\,\mathrm{bwd}}(v;\,G_{<i},\tau),
  \label{eq:retrieve}
\end{equation}
where $\mathrm{Expand}$ collects the nodes reachable within $h$ backward hops.
The backward-only direction over
$G_{<i}$ enforces causality at the graph level, while $\tau$ prunes weak
associations so expansion does not drown the relevant context. The expansion
recovers entities absent from the immediate neighbors yet continuous over a
longer range (Figure~\ref{fig:edges}(b)); with no explicit edges it falls back to
implicit links induced on the fly. Our lexical scoring follows the probabilistic
relevance tradition of BM25~\cite{bm25_2009} but uses no embeddings, so retrieval
adds no model dependency.

Not every retrieved candidate should be injected. Indiscriminate concatenation of
$\mathcal{K}_i$ would overwrite legitimate creative change and pollute the prompt
with redundant or conflicting context. We therefore apply a \emph{selective
filtering} step that judges each candidate independently. For a candidate
$c\in\mathcal{K}_i$, a decision function returns a keep/discard indicator together
with a justification,
\begin{equation}
  \phi(c,s_i)\in\{0,1\},\qquad
  \mathcal{K}_i^{\ast}=\{\,c\in\mathcal{K}_i \mid \phi(c,s_i)=1\,\},
  \label{eq:filter}
\end{equation}
where $\phi$ is realized in the online mode by an LLM that scores candidates
against $s_i$ under an explicit rubric, discarding one that stands on its own, is
redundant, or is separated from $s_i$ by a scene cut, and keeping one only when it
supplies genuine cross-shot continuity, with an auditable justification. In the
offline mode $\phi$ degrades to a conservative rule that keeps a candidate
whenever the shot has any prior context. Retaining only $\mathcal{K}_i^{\ast}$ is
what makes the memory \emph{selective}.

\subsubsection{Natural-Rewrite Injection}
\label{sec:method_inject}
Given the filtered context $\mathcal{K}_i^{\ast}$, an injection operator $g$
produces the continuity-enriched prompt
\begin{equation}
  p_i=g(s_i,\mathcal{K}_i^{\ast}),
  \label{eq:inject}
\end{equation}
which replaces the stateless $f_\theta(s_i)$ of Eq.~\eqref{eq:baseline}. We
realize $g$ in two modes. The preferred
mode is a \emph{natural rewrite}: an LLM weaves the retained continuity into the
visual description $x_i$ in the director's own register, adding only the
continuity constraints while leaving the shot grammar $\kappa_i$ and the visual
semantics unchanged. Because an unconstrained rewrite could silently drift from the
director's intent, we accept the rewritten description $x_i'$ only when it passes
a guard $\mathcal{V}(x_i',x_i)$ that (i) bounds the length ratio
$|x_i'|/|x_i|\in[\rho_{\min},\rho_{\max}]$, (ii) preserves every resource
reference of $x_i$, and (iii) keeps its structural markers; a rewrite failing the
guard is rejected and the shot reverts to its previous description. As a
lightweight fallback, a \emph{mechanical injection} mode appends
$\mathcal{K}_i^{\ast}$ at a fixed prompt position without an LLM. Both modes add
continuity rather than rewriting intent, and the guard preserves the director's
original creative decisions.

\subsection{\pipename: \sysname Within a Multi-Agent Pipeline}
\label{sec:method_pipeline}

\begin{figure*}[t]
  \centering
  \includegraphics[width=\textwidth]{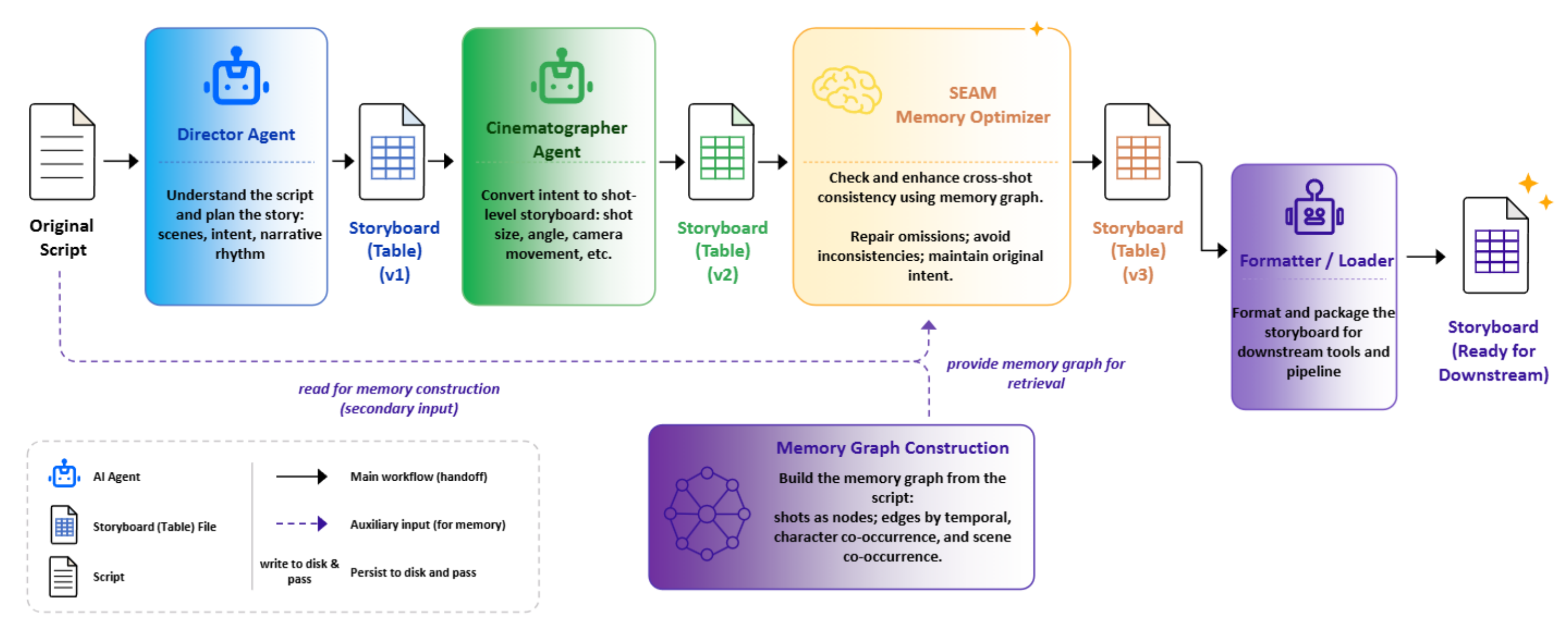}
  \Description{Left-to-right pipeline diagram: Director Agent, Cinematographer
    Agent, SEAM Memory Optimizer, and Formatter, with the memory optimizer
    placed between the authoring agents and the downstream loader.}
  \caption{\sysname sits as the ``memory optimizer'' stage, placed after the
    authoring agents and before the loader. It builds the memory graph from the
    episode script and repairs continuity shot by shot over the shotlist that the
    authoring agents have already generated.}
  \label{fig:pipeline}
\end{figure*}

In this section, we describe how \sysname is deployed as one stage of \pipename.
As shown in Figure~\ref{fig:pipeline}, \pipename is a serial multi-agent pipeline
in which the agents communicate through shotlist artifacts on disk, and \sysname
operates as a memory optimizer between authoring and rendering.

\subsubsection{Pipeline Structure}
\label{sec:method_pipeline_struct}
\pipename comprises four serial stages. A \textbf{Director Agent} reads the
episode script and drafts the overall scene intent and narrative beats; a
\textbf{Cinematographer Agent} turns that intent into a concrete shotlist,
assigning the shot grammar $\kappa_i$ (shot size, angle, and movement) to each
shot; the \textbf{\sysname Memory Optimizer} performs cross-shot continuity
repair over that shotlist; and a \textbf{Formatter} standardizes the result into
the tabular schema that downstream tools consume. Each stage writes an artifact
that the next stage reads, so the chain is decoupled at the file level. \sysname
is a mandatory stage in this chain, not an optional pass that runs afterward, so
every shotlist the pipeline emits has been continuity-checked before the
downstream keyframe and video generators consume it.

\subsubsection{\sysname as the Memory-Optimizer Stage}
\label{sec:method_pipeline_agent}
The memory optimizer takes two inputs: the episode script, from which it builds
and persists the memory graph, and the upstream shotlist, whose descriptions it
repairs. Applying the three-stage procedure of
Section~\ref{sec:method_framework}, it extracts the shot states into $G$,
retrieves and filters each shot's prior context over $G_{<i}$, and rewrites the
description via Eq.~\eqref{eq:inject}, emitting $p_i=g(s_i,\mathcal{K}_i^{\ast})$
in place of the stateless baseline $f_\theta(s_i)$ of Eq.~\eqref{eq:baseline}.
Because state accumulates across episodes, the retrieved context spans earlier
shots of the current episode and prior episodes alike, which is what lets the
optimizer repair the cross-episode breaks that dominate long titles. Since it
operates solely at the prompt-text layer, it is decoupled from both the authoring
model and the downstream backbone, a property we verify with the six-model results
of Section~\ref{sec:exp}.

\section{Experiments}
\label{sec:exp}

We study three research questions. \textbf{Q1}: does the memory graph repair the
cross-shot and cross-episode continuity defects a stateless pipeline leaves
behind? \textbf{Q2}: is this repair model-agnostic across heterogeneous
backbones? \textbf{Q3}: do prompt-layer repairs survive the text-to-image stage
and stay measurable in the keyframes?

\subsection{Experimental Setup}
\label{sec:exp_bench}

\paragraph{\benchname.}
As summarized in Table~\ref{tab:bench}, \benchname is a short-drama continuity
storyboarding benchmark over three produced dramas totaling 68 episodes, referred
to throughout by the
short identifiers of the released data: \texttt{his-toyboy} (\textit{His Toyboy:
The Billionaire's Trap}), \texttt{beyond-the-wall} (\textit{Beyond the Wall}), and
\texttt{werewolf} (\textit{You Are My Cure, My Undoing}). We release two kinds of
material: the original scripts forming the pipeline input $s_i$, and reference
images---character/scene visual anchors from a semantically renamed, deduplicated
pool---serving as fixed keyframe input and as the image-layer gold standard. Expert human-director
storyboards annotated shot by shot are the professional reference,
but remain restricted by copyright, held out from the release, and used only for
evaluation and unblinding.

\begin{table}[t]
  \centering
  \small
  \caption{\benchname data statistics. AI shots are counted on the
    \texttt{claude} run; other backbones differ by $<5\%$.}
  \label{tab:bench}
  \scalebox{\tabscalebench}{%
  \begin{tabular}{@{}lrrrr@{}}
    \toprule
    Drama & Ep. & Human & AI & Ref.\ img. \\
    \midrule
    \texttt{his-toyboy}      & 23 & 754   & 907   & 15 \\
    \texttt{beyond-the-wall} & 20 & 659   & 814   & 31 \\
    \texttt{werewolf}        & 25 & 826   & 997   & 12 \\
    \midrule
    Total           & 68 & 2{,}239 & 2{,}718 & 58 \\
    \bottomrule
  \end{tabular}}
\end{table}

\paragraph{Setup.}
We evaluate six heterogeneous text models (\texttt{claude},
\texttt{deepseek-v4-pro}, \texttt{glm-5.1}, \texttt{gpt-5.4},
\texttt{kimi-k2.6}, \texttt{minimax-m2.7}). Per episode, each backbone produces
an uninjected \emph{shotlist1} (Eq.~\eqref{eq:baseline}) and a memory-injected
\emph{shotlist2} (Eq.~\eqref{eq:inject}). The two CSVs are column- and shot-aligned,
making every comparison a paired within-model contrast whose only varying factor
is memory injection. Keyframes come from Nano Banana Pro on both variants under
identical references and template skeletons, from which we extract 900 test
instances for the image layer, of which 820 returned an image. Scoring is delegated to
Gemini~3 Pro, excluded from the six evaluated backbones so that no model judges its
own storyboards.

\paragraph{Evaluation protocol.}
Storyboarding has no established automatic text-layer metric, and recent
script-to-screen systems rely on human Likert ratings or LLM judges
alone~\cite{movieagent2025,directorllm2024}. To keep no conclusion resting on a
single evaluator family, we pair a double-blind LLM judge with a CPU-only offline
metric suite assembled from adjacent
literatures~\cite{sbert2019,ceaf2005,bertscore2020,lstoryboard2025,et2024,entitygraph2013,summac2022,mdeberta_xnli2022,cmlbench2025}.
The suite scores a storyboard along four complementary axes: agreement with the
human director, accuracy of the cinematography fields, camera-label distribution,
and reference-free continuity. An image layer then scores the rendered keyframes
against their
references~\cite{dreambooth2023,consistory2024,vistorybench2025,dinov2_2024,videomemory2026,dreamsim2023,vbench2024,clip2021,laion_aesthetic2022}.
Definitions and formulas are deferred to Appendix~\ref{app:metrics}, where
Table~\ref{tab:abbrev} lists every abbreviation used below. In the double-blind
judge, conditions appear as ``A/B'' with sources hidden and deterministically
counterbalanced, revealed only after scoring; the prompt layer is scored on five
dimensions and the image layer on four. Every baseline-vs-memory contrast is
paired at the episode level and tested with the Wilcoxon signed-rank test under
Holm correction~\cite{wilcoxon1945,holm1979} ($^{*}$: corrected $p<0.05$). The 80
renders refused by the provider's safety filter (triggered by sensitive plot
content, near-symmetric across conditions: 412 baseline vs.\ 408 memory images
survive) are dropped pairwise, so a shot enters the image metrics only if both
conditions rendered it, leaving 403 paired shots. The NLI and image layers run on
\texttt{beyond-the-wall}, the rest on all three dramas.

\subsection{Main Results (Q1/Q2/Q3)}
\label{sec:exp_evidence}

\paragraph{Q1: memory injection repairs continuity.}
Three measurements triangulate the repair effect. First, we turn to
\emph{continuity recall} (Eq.~\eqref{eq:recall}). As shown in
Table~\ref{tab:h1}, on \texttt{his-toyboy} (23 episodes) the memory
graph repairs $35$ of the $37$ defects the judge rules in need of repair
($0.946$), against $7$ of $10$ ($0.700$) for a control whose retrieval is
restricted to the current episode. The two conditions do not
share a denominator, and the reason is itself part of the result: a defect can
only be judged once retrieval surfaces the prior state it contradicts, so the
episode-local control yields $3.7\times$ fewer candidate sites. \sysname
therefore improves on two axes at once, exposing $27$ additional genuine defects
and repairing a larger fraction of those it exposes; the recall column understates
the gap, since those $27$ are left unrepaired rather than counted against the
control. Resting on different defect populations, this measurement establishes
that cross-episode retrieval is what makes defects addressable, and we rely on
the next two, both computed per shot, for the magnitude of the repair. Second, the
\emph{targeted state-conditioned contradiction} probe scores each filter-flagged
shot (Section~\ref{sec:method_retrieve}) against its graph-retrieved expected
state as NLI premise, before versus after injection. Table~\ref{tab:targeted}
reports the outcome: across all 2{,}946 flagged shots, mean contradiction
probability drops from $0.308$ to $0.131$ and the threshold-exceeding fraction
from $0.264$ to $0.093$, removing roughly two thirds of the contradiction mass at
the targeted sites in every drama--backbone cell, with a residual 4--7\% resisting
repair that Figure~\ref{fig:targeted_scatter} exposes and
Section~\ref{sec:exp_analysis} takes up. Third, the
\emph{double-blind prompt-layer judgment} over six backbones (30 model-episodes,
episodes 1--5 of \texttt{beyond-the-wall}, the same subset carrying the image
layer) favors memory on all five dimensions. As Table~\ref{tab:blind} and
Figure~\ref{fig:blind_dumbbell} show, character-state continuity ($+4.4$,
$r{=}0.82$), prop continuity ($+2.9$, $r{=}0.77$), and intent fit ($+6.5$,
$r{=}0.90$) stay significant after Holm correction.

\begin{table}[t]
  \centering
  \small
  \caption{Q1 continuity recall (\texttt{his-toyboy}, 23 episodes). The control
    restricts retrieval to the current episode, so it surfaces fewer judgeable
    defects; \#Judged is therefore part of the effect rather than a fixed
    denominator (see text).}
  \label{tab:h1}
  \scalebox{\tabscalehone}{%
  \begin{tabular}{@{}lccc@{}}
    \toprule
    Condition & Recall$\uparrow$ & \#Judged & \#Modified \\
    \midrule
    Control (episode-local retrieval) & 0.700 & 10 & 7  \\
    \sysname (cross-episode graph)    & 0.946 & 37 & 35 \\
    \bottomrule
  \end{tabular}}
\end{table}

\begin{table}[t]
  \centering
  \small
  \caption{Targeted state-conditioned contradiction on repair-flagged shots.
    The premise is the graph-expected entity state and the score is the NLI
    contradiction probability before vs.\ after injection, pooled over six
    backbones per drama and weighted by flagged-shot count.}
  \label{tab:targeted}
  \setlength{\tabcolsep}{5pt}
  \scalebox{\tabscaletargeted}{%
  \begin{tabular}{@{}lrcccc@{}}
    \toprule
    & & \multicolumn{2}{c}{Mean prob.$\downarrow$} &
      \multicolumn{2}{c}{Rate ($p{\ge}0.5$)$\downarrow$} \\
    \cmidrule(lr){3-4}\cmidrule(lr){5-6}
    Drama & \#Flag. & before & after & before & after \\
    \midrule
    \texttt{beyond-the-wall} & 1{,}085 & 0.344 & 0.106 & 0.320 & 0.069 \\
    \texttt{his-toyboy}      & 923     & 0.302 & 0.157 & 0.255 & 0.122 \\
    \texttt{werewolf}        & 938     & 0.273 & 0.133 & 0.209 & 0.091 \\
    \midrule
    All             & 2{,}946 & 0.308 & 0.131 & 0.264 & 0.093 \\
    \bottomrule
  \end{tabular}}
\end{table}

\paragraph{Q2: model-agnosticism.}
Table~\ref{tab:h2} consolidates the per-model evidence across the three metric
families. Recall (panel~a) is high in all 18 drama--backbone cells, from $0.667$
(\texttt{claude} on \texttt{his-toyboy}) to $1.000$ (\texttt{minimax} on
\texttt{beyond-the-wall}), with per-model means spanning $0.731$--$0.992$. The
targeted contradiction drop (panel~b) holds throughout, mean probability falling
from the $0.29$--$0.35$ band to $0.10$--$0.17$. The blind gains (panel~c) are
positive on 17 of 18 model--dimension entries; the sole exception, prop continuity
for \texttt{glm-5.1}, is a $-0.04$ shift rounding to zero. The spread reflects each
model's style, yet the direction never reverses, evidence of model-agnosticism.

The ranking is not an artifact of defect volume. Judged defects per episode vary
by an order of magnitude across cells ($1.7$--$19.8$, or $3.6$--$13.7$ per model),
yet five of six per-model means land in a tight $0.93$--$0.99$ band.
\texttt{claude} is the sole outlier, lowest on recall not because it judges more
defects (it flags the \emph{fewest}) but because its rewrites are the most
conservative, and recall counts a defect repaired only when a changed description
is emitted; on the blind judgment it still posts among the largest character-state
gains ($+5.1$). Since recall cannot separate ``no rewrite'' from ``wrong
rewrite,'' we triangulate with the targeted probe and the blind judge.

\begin{table}[t]
  \centering
  \small
  \caption{Per-model results over the six backbones in three metric groups:
    \emph{(a)} continuity recall per drama---\textsc{btw}
    (\texttt{beyond-the-wall}), \textsc{hty} (\texttt{his-toyboy}), \textsc{wlf}
    (\texttt{werewolf})---and its mean; \emph{(b)} targeted contradiction
    probability before/after injection, pooled over dramas and weighted by
    flagged-shot count ($n$); \emph{(c)} blind prompt-layer gains
    $\Delta{=}\text{M}{-}\text{B}$ on the memory-edited dimensions
    (Table~\ref{tab:abbrev}). Every cell improves, answering Q2 in the affirmative.}
  \label{tab:h2}
  \setlength{\tabcolsep}{2.2pt}
  \setlength{\aboverulesep}{0pt}\setlength{\belowrulesep}{0pt}
  \renewcommand{\arraystretch}{0.92}
  \scalebox{\tabscalehtwo}{%
  \begin{tabular}{@{}lcccccccccc@{}}
    \toprule
    & \multicolumn{4}{c}{\emph{(a) Recall}$\uparrow$}
      & \multicolumn{3}{c}{\emph{(b) Contradiction}}
      & \multicolumn{3}{c}{\emph{(c) $\Delta$Blind}$\uparrow$} \\
    \cmidrule(lr){2-5}\cmidrule(lr){6-8}\cmidrule(lr){9-11}
    Backbone & \textsc{btw} & \textsc{hty} & \textsc{wlf} & \textbf{mean}
      & $n$ & bef.$\downarrow$ & aft.$\downarrow$
      & \textsc{csc} & \textsc{prp} & \textsc{ifit} \\
    \midrule
    \texttt{claude}   & 0.760 & 0.667 & 0.767 & \textbf{0.731}
      & 178 & 0.321 & 0.129 & $+5.1$ & $+0.9$ & $+4.9$ \\
    \texttt{deepseek} & 0.912 & 0.959 & 0.915 & \textbf{0.929}
      & 863 & 0.286 & 0.115 & $+9.9$ & $+5.9$ & $+15.8$ \\
    \texttt{glm-5.1}  & 0.983 & 0.947 & 0.960 & \textbf{0.963}
      & 248 & 0.308 & 0.116 & $+3.5$ & $-0.0$ & $+2.8$ \\
    \texttt{gpt-5.4}  & 0.994 & 0.988 & 0.995 & \textbf{0.992}
      & 548 & 0.294 & 0.135 & $+2.1$ & $+1.1$ & $+2.8$ \\
    \texttt{kimi}     & 0.961 & 0.928 & 0.960 & \textbf{0.950}
      & 399 & 0.353 & 0.103 & $+2.7$ & $+1.0$ & $+7.6$ \\
    \texttt{minimax}  & 1.000 & 0.967 & 0.993 & \textbf{0.987}
      & 710 & 0.319 & 0.167 & $+3.0$ & $+8.8$ & $+4.8$ \\
    \bottomrule
  \end{tabular}}
\end{table}

\paragraph{Q3: transfer to the image layer.}
As Table~\ref{tab:blind}b shows, at the generated-image layer the blind judge
still prefers memory on the two dimensions injection actually edits, character
appearance ($+3.7$) and prop continuity ($+3.3$), while scene layout and frame
quality stay flat. With only $n{=}30$ sequence pairs these deltas do not
reach significance, so we answer Q3 only directionally: transfer is visible but
attenuated by the stochastic text-to-image stage.

\begin{table}[t]
  \centering
  \small
  \caption{Double-blind LLM judgment (0--100) per dimension, pooled over six
    backbones (\texttt{beyond-the-wall}, ep.\ 1--5): \emph{B} baseline, \emph{M}
    memory, $\Delta{=}\text{M}{-}\text{B}$ ($^{*}$ Holm-corrected $p<0.05$);
    $n{=}30$ model-episodes and $30$ sequences.}
  \label{tab:blind}
  \setlength{\tabcolsep}{10pt}
  \scalebox{\tabscaleblind}{%
  \begin{tabular}{@{}lccccc@{}}
    \toprule
    \multicolumn{6}{@{}l}{\emph{(a) Prompt layer (per-shot)}} \\
    \midrule
    & \textsc{csc}$\uparrow$ & \textsc{scon}$\uparrow$ & \textsc{prp}$\uparrow$
      & \textsc{cgen}$\uparrow$ & \textsc{ifit}$\uparrow$ \\
    \cmidrule(lr){2-6}
    B        & 88.1 & 91.8 & 92.7 & 91.1 & 87.8 \\
    M        & 92.5 & 92.9 & 95.7 & 91.7 & 94.2 \\
    $\Delta$ & $+4.4^{*}$ & $+1.1$ & $+2.9^{*}$ & $+0.6$ & $+6.5^{*}$ \\
    \midrule
    \multicolumn{6}{@{}l}{\emph{(b) Image layer (per-sequence)}} \\
    \midrule
    & \textsc{capp}$\uparrow$ & \textsc{slay}$\uparrow$ & \textsc{prpi}$\uparrow$
      & \textsc{sfq}$\uparrow$ & \\
    \cmidrule(lr){2-5}
    B        & 64.0 & 84.3 & 62.3 & 93.0 & \\
    M        & 67.7 & 83.1 & 65.7 & 94.0 & \\
    $\Delta$ & $+3.7$ & $-1.2$ & $+3.3$ & $+1.0$ & \\
    \bottomrule
  \end{tabular}}
\end{table}

\paragraph{Offline metrics: what moves and what does not.}
Table~\ref{tab:offline} shows how the offline suite separates \emph{targeted
repair} from \emph{global drift}. Against the human director the paired shifts sit
in the third decimal and the divergences stay at zero, since injection edits
entity-state phrases without touching the cinematography fields, while the
reference-free layer moves in the expected direction on every axis; we return to
this contrast in Section~\ref{sec:exp_analysis}, where the two-thirds targeted
drop quantifies it. The offline image embeddings show the same attenuation, all
six scores flat at this sample size, consistent with generation noise dominating.

\begin{table}[t]
  \centering
  \small
  \caption{Offline text-layer metrics over 408 model-episodes, reference-based
    (top) and reference-free (bottom): \emph{s1} baseline, \emph{s2} memory,
    $\Delta{=}\text{s2}{-}\text{s1}$ ($^{*}$ Holm-corrected $p<0.05$);
    $\uparrow$/$\downarrow$ is the improvement direction. \textsc{js} pools the
    three divergences, bit-identical across conditions; NLI runs on
    \texttt{beyond-the-wall} ($n{=}192$).}
  \label{tab:offline}
  \setlength{\tabcolsep}{4.5pt}
  \scalebox{\tabscaleoffline}{%
  \begin{tabular}{@{}lccccc@{}}
    \toprule
    & \textsc{sm-f1}$\uparrow$ & \textsc{bs-f1}$\uparrow$
      & \textsc{sz-acc}$\uparrow$ & \textsc{k$\tau$}$\uparrow$
      & \textsc{js}$\downarrow$ \\
    \cmidrule(lr){2-6}
    s1       & 0.4755 & 0.6379 & 0.318 & 0.446 & fixed \\
    s2       & 0.4731 & 0.6367 & 0.313 & 0.412 & fixed \\
    $\Delta$ & $-0.002^{*}$ & $-0.001^{*}$ & $-0.005$ & $-0.034^{*}$ & $0.000$ \\
    \midrule
    & \textsc{eod}$\uparrow$ & \textsc{acos-m}$\uparrow$
      & \textsc{acos-n}$\uparrow$ & \textsc{nli-cr}$\downarrow$
      & \textsc{nli-mp}$\downarrow$ \\
    \cmidrule(lr){2-6}
    s1       & 11.06 & 0.4860 & 0.148 & 0.668 & 0.656 \\
    s2       & 11.06 & 0.4908 & 0.148 & 0.660 & 0.649 \\
    $\Delta$ & $0.000$ & $+0.005^{*}$ & $0.000$ & $-0.009^{*}$ & $-0.007^{*}$ \\
    \bottomrule
  \end{tabular}}
\end{table}

\subsection{Online Deployment Evidence for Memory Effectiveness}
\label{sec:exp_online}

Beyond the offline suite, we report a live online run of \sysname as the mandatory
memory-optimization stage of \pipename in CreativeFitting's production system, over
a freshly produced short-drama of four consecutive episodes (201 shots), whose
deployment architecture is detailed in Appendix~\ref{app:deployment}. Human
annotators decide acceptance under a human-specified four-state standard, a proxy
for \emph{director acceptance}.

As summarized in Table~\ref{tab:online_adoption}, \sysname reached an overall
director-acceptance rate of $96.5\%$ at a zero
unsafe-injection error rate, counting an injection unsafe if it contradicts the
shot it edits or overwrites the director's stated intent. The residual misses share one failure mode: a persistent
state is tracked correctly in some shots of a run but not all, so the injection is
incomplete rather than wrong. To bound how much of the rate the memory explains, we
compute a pessimistic counterfactual: reclassifying every accepted shot whose
repair drew on a prior episode as a miss drops acceptance to $74.6\%$, a lower
bound of $\Delta=21.9$\,pp on the contribution of cross-episode memory. It is
conservative: it discards such a shot outright instead of crediting the
episode-local continuity it would still have received. The mechanism is restrained:
it injects into only $27.4\%$ of shots, with minimal fragments (median ratio
$27.6\%$) and non-injected shots unchanged to the byte, so memory adds continuity
without rewriting intent, corroborating Q1 in production. These storyboards carry
commercially released work: titles from this pipeline reach viewers on Reel.AI,
where \textit{Lost Before I Found You} has topped the DataEye overseas micro-drama
heat ranking by a wide margin over the live-action titles that otherwise dominate
it. Chart position is no controlled measurement and we claim no causal credit, but
it does place the $96.5\%$ on production work, not on a pilot.

\begin{table}[t]
  \centering
  \small
  \caption{Online memory-effectiveness over the four deployed episodes. The
    counterfactual row is a \emph{conservative lower bound}: every
    cross-episode-dependent repair reclassified as a miss.}
  \label{tab:online_adoption}
  \setlength{\tabcolsep}{6pt}
  \scalebox{\tabscaleadoption}{%
  \begin{tabular}{@{}lr@{}}
    \toprule
    Metric & Value \\
    \midrule
    Shots evaluated                                          & 201 \\
    Director-acceptance rate$\uparrow$                       & 96.5\% \\
    \quad w/o cross-episode memory (lower bound)$\uparrow$   & 74.6\% \\
    \quad memory attribution ($\Delta$)$\uparrow$            & $+21.9$\,pp \\
    Selective memory-injection rate                          & 27.4\% \\
    Injection edit locality (median)                         & 27.6\% \\
    Unsafe-injection error rate$\downarrow$                  & 0.0\% \\
    \bottomrule
  \end{tabular}}
\end{table}

\subsection{Further Analysis}
\label{sec:exp_analysis}

\paragraph{Targeted repair versus global dilution.}
As shown in Figure~\ref{fig:dilution}, which puts the two measurement scopes on a
common relative scale, the contradiction rate and probability improve by
$64.9\%$ and $57.7\%$ at the flagged sites, whereas every episode-level metric
moves by at most $1.3\%$.
The ratio indicates a sparse effect, not a weak one: the filter flags roughly one
shot in ten, so an episode-level average spreads a near-total repair over an order
of magnitude more unaffected shots, which is why we measure continuity at flagged
sites and why Table~\ref{tab:offline} stays flat. Nor is the drop carried by a
favorable subset: Figure~\ref{fig:targeted_heatmap} resolves
Table~\ref{tab:targeted} into all 18 drama--backbone cells, and the mean
probability falls in every one, from $0.25$--$0.39$ to $0.05$--$0.20$, so the
repair floor is set by the memory graph, not the authoring model.

\paragraph{Failure modes and limitations.}
We report the failures directly. \emph{(i)} Not every judged defect is repaired:
37 judged versus 35 modified for Q1, and $4$--$7\%$ of flagged shots stay above
threshold, which Figure~\ref{fig:targeted_scatter} makes visible as the mass above
the diagonal. \emph{(ii)} Image-layer transfer, as Table~\ref{tab:blind}b reports,
is directional but not significant at $n{=}30$, so Q3 rests on the weakest
evidence of the three. \emph{(iii)} Kendall's
$\tau$ drops under injection ($0.446$ to $0.412$, $p<0.05$), a reordering side
effect the rewrite guard does not constrain. \emph{(iv)} Camera-angle accuracy
stays near $0.20$ in both conditions, so the shot-grammar gap remains open. Two
limits further bound the evidence: one judge family supplies both the recall labels
and the blind scores, mitigated but not removed by the offline suite; and the
fallback modes of Section~\ref{sec:method_framework} are unevaluated.

\section{Conclusion}
\label{sec:conclusion}

We presented \sysname, a training-free, model-agnostic memory graph that repairs
cross-shot and cross-episode continuity in prompt text. It lifts continuity
recall from $0.700$ to $0.946$, stays positive across six text models
($0.731$--$0.992$), and transfers directionally to images. \benchname is
released, and \sysname runs as a mandatory \pipename stage in production. Future
work targets video-layer evaluation.

\bibliographystyle{ACM-Reference-Format}
\bibliography{cfgraph}

\appendix
\newcommand{\mlayer}[1]{\par\addvspace{\medskipamount}%
  \noindent\textbf{\textit{#1}}\quad\ignorespaces}
\clearpage
\section{Metric Definitions}
\label{app:metrics}

This appendix gives the full definition, formula, and computation of every
metric used in Section~\ref{sec:exp}. Table~\ref{tab:abbrev} is the master
abbreviation list; the abbreviations are used throughout the experiment
tables. Unless stated otherwise, all embedding-based metrics use the
multilingual sentence encoder
\texttt{paraphrase-multilingual-MiniLM-L12-v2} with $L_2$-normalized outputs,
so that a dot product equals a cosine similarity. This encoder is chosen
because AI prompts are in English while human-director descriptions are in
Chinese, requiring cross-lingual alignment. Every metric runs on CPU.

\begin{table}[t]
  \centering
  \small
  \caption{Master list of metric abbreviations. ``Dir.'' is the improvement
    direction: $\uparrow$ higher is better, $\downarrow$ lower is better.}
  \label{tab:abbrev}
  \setlength{\tabcolsep}{4pt}
  \scalebox{\tabscaleabbrev}{%
  \begin{tabular}{@{}llc@{}}
    \toprule
    Abbr. & Full name & Dir. \\
    \midrule
    \multicolumn{3}{@{}l}{\emph{Alignment layer (reference-based)}} \\
    \textsc{sm-p}    & Soft-Match Precision            & $\uparrow$ \\
    \textsc{sm-r}    & Soft-Match Recall               & $\uparrow$ \\
    \textsc{sm-f1}   & Soft-Match F1                   & $\uparrow$ \\
    \textsc{bs-f1}   & BERTScore F1                    & $\uparrow$ \\
    \textsc{k$\tau$} & Kendall's $\tau$ (ordering)     & $\uparrow$ \\
    \midrule
    \multicolumn{3}{@{}l}{\emph{Field layer (reference-based)}} \\
    \textsc{sz-acc}  & Shot-size Accuracy              & $\uparrow$ \\
    \textsc{sz-f1}   & Shot-size macro-F1              & $\uparrow$ \\
    \textsc{an-acc}  & Camera-angle Accuracy           & $\uparrow$ \\
    \textsc{an-f1}   & Camera-angle macro-F1           & $\uparrow$ \\
    \textsc{mv-jac}  & Movement Jaccard (multi-label)  & $\uparrow$ \\
    \midrule
    \multicolumn{3}{@{}l}{\emph{Distribution layer (reference-based)}} \\
    \textsc{sz-js}   & Shot-size distribution JS div.\ & $\downarrow$ \\
    \textsc{an-js}   & Camera-angle distribution JS div.\ & $\downarrow$ \\
    \textsc{mv-js}   & Movement distribution JS div.\  & $\downarrow$ \\
    \midrule
    \multicolumn{3}{@{}l}{\emph{Reference-free layer}} \\
    \textsc{eod}     & Entity Out-Degree               & $\uparrow$ \\
    \textsc{acos-m}  & Adjacent Cosine (mean)          & $\uparrow$ \\
    \textsc{acos-n}  & Adjacent Cosine (min)           & $\uparrow$ \\
    \textsc{nli-cr}  & NLI Contradiction Rate          & $\downarrow$ \\
    \textsc{nli-mp}  & NLI Mean max contradiction Prob.\ & $\downarrow$ \\
    \midrule
    \multicolumn{3}{@{}l}{\emph{Targeted probe (state-conditioned)}} \\
    \textsc{tc-p}    & Targeted Contradiction Prob.\   & $\downarrow$ \\
    \textsc{tc-r}    & Targeted Contradiction Rate     & $\downarrow$ \\
    \midrule
    \multicolumn{3}{@{}l}{\emph{Image layer}} \\
    \textsc{cref}    & Character--Reference DINOv2 sim.\ & $\uparrow$ \\
    \textsc{cself}   & Character self-consistency (DreamSim dist.)\ & $\downarrow$ \\
    \textsc{sref}    & Scene--Reference DINOv2 sim.\   & $\uparrow$ \\
    \textsc{sself}   & Scene self-consistency DINOv2 sim.\ & $\uparrow$ \\
    \textsc{clip-t}  & CLIP-T prompt fidelity          & $\uparrow$ \\
    \textsc{aes}     & LAION Aesthetic score           & $\uparrow$ \\
    \midrule
    \multicolumn{3}{@{}l}{\emph{Double-blind judge, prompt layer}} \\
    \textsc{csc}     & Character-State Continuity      & $\uparrow$ \\
    \textsc{scon}    & Scene Consistency               & $\uparrow$ \\
    \textsc{prp}     & Prop Continuity                 & $\uparrow$ \\
    \textsc{cgen}    & Composition Generability        & $\uparrow$ \\
    \textsc{ifit}    & Intent Fit                      & $\uparrow$ \\
    \midrule
    \multicolumn{3}{@{}l}{\emph{Double-blind judge, image layer}} \\
    \textsc{capp}    & Character Appearance            & $\uparrow$ \\
    \textsc{slay}    & Scene Layout                    & $\uparrow$ \\
    \textsc{prpi}    & Prop Continuity (image)         & $\uparrow$ \\
    \textsc{sfq}     & Single-Frame Quality            & $\uparrow$ \\
    \midrule
    \multicolumn{3}{@{}l}{\emph{Continuity recall (LLM-judged)}} \\
    \textsc{rec}     & Continuity Recall               & $\uparrow$ \\
    \bottomrule
  \end{tabular}}
\end{table}

\begin{table}[t]
  \centering
  \small
  \caption{Multi-dimensional definition of shot state ($d{=}8$ dimensions),
    referenced from Section~\ref{sec:method_extract}.}
  \label{tab:state_dims}
  \scalebox{\tabscalestatedims}{%
  \begin{tabular}{@{}ll@{}}
    \toprule
    Dimension & Meaning \\
    \midrule
    scene            & physical scene / location \\
    atmosphere       & lighting, color tone, mood \\
    characters       & on-screen characters \\
    character states & per-character emotion and posture \\
    spatial          & spatial relations / blocking of characters, objects \\
    props            & key props and possession relations \\
    actions          & in-shot actions / events \\
    camera style     & shot size / angle / movement \\
    \bottomrule
  \end{tabular}}
\end{table}

\mlayer{Continuity recall (\textsc{rec}).}
The core continuity metric. For an entity $e$ and a shot pair $i<j$ in which
$e$ is active in both, a continuity defect $\delta(e,i,j){=}1$ is recorded when
$\pi_i(e)$ contradicts $\pi_j(e)$ without narrative motivation
(Section~\ref{sec:prelim}). The LLM judge labels which defects need repair
and which repairs succeed, giving Eq.~\eqref{eq:recall}:
\begin{equation*}
  \textsc{rec}=
    \frac{\#\{\text{defects successfully repaired}\}}
         {\#\{\text{defects judged in need of repair}\}}\in[0,1].
\end{equation*}

\mlayer{Alignment layer (reference-based).}
AI and human-director storyboards segment the same script at different
granularities, so we soft-align them before any field comparison. The
per-shot visual descriptions of both sides are encoded and form a cosine
similarity matrix $\mathrm{sim}\in\mathbb{R}^{N_{\mathrm{AI}}\times
M_{\mathrm{human}}}$; a Hungarian assignment on $-\mathrm{sim}$ yields the
optimal one-to-one matching, and pairs with similarity below $\tau_{\min}{=}0.3$
are discarded (CEAF-style~\cite{ceaf2005}).

\paragraph{Soft-match P/R/F1 (\textsc{sm-p/r/f1}).}
Let $\Phi^{*}$ be the total similarity mass of the retained matched pairs,
$N_{\mathrm{AI}}$ the AI shot count, and $M_{\mathrm{human}}$ the
human-director shot count. Then
\begin{equation}
  \textsc{sm-p}=\frac{\Phi^{*}}{N_{\mathrm{AI}}},\quad
  \textsc{sm-r}=\frac{\Phi^{*}}{M_{\mathrm{human}}},\quad
  \textsc{sm-f1}=\frac{2\,\textsc{sm-p}\cdot\textsc{sm-r}}
                      {\textsc{sm-p}+\textsc{sm-r}}.
  \label{eq:softmatch}
\end{equation}
These measure how well the AI storyboard matches the director's in both
semantics and shot count.

\paragraph{BERTScore F1 (\textsc{bs-f1}).}
For every matched description pair we compute a token-level BERTScore
with the multilingual \texttt{bert-base-multilingual-cased} checkpoint and
report the mean $F_1$ over pairs. It captures matched-pair description
quality at a finer granularity than the sentence-level cosine.

\paragraph{Kendall's $\tau$ (\textsc{k$\tau$}).}
Sorting matched pairs by their AI-side index and reading off the
reference-side indices gives a permutation; with $C$ concordant and $D$
discordant pairs over $n$ matched shots,
\begin{equation*}
  \textsc{k$\tau$}=\frac{C-D}{n(n-1)/2}\in[-1,1],
\end{equation*}
measuring how well the AI shot ordering preserves the director's ordering.

\mlayer{Field layer (reference-based).}
On the matched pairs we compare three cinematography fields, each normalized
by a regex rule table into a closed label set: shot size (7 classes), camera
angle (9 classes), and camera movement (12 classes, multi-label). Unmatched
raw values map to \emph{unknown} and each field additionally reports its
coverage (comparable pairs / matched pairs).

\paragraph{Accuracy and macro-F1 (\textsc{sz/an-acc}, \textsc{sz/an-f1}).}
For the single-label size and angle fields, accuracy is the fraction of
matched pairs with $\mathrm{pred}{=}\mathrm{gold}$, and macro-F1 is the
unweighted mean of per-class $F_1$ over classes appearing in gold or pred.

\paragraph{Movement Jaccard (\textsc{mv-jac}).}
Movement is multi-label; for each matched pair with label sets $A$ (pred) and
$B$ (gold),
\begin{equation*}
  \textsc{mv-jac}=\frac{1}{|\mathcal{M}|}\sum_{(A,B)\in\mathcal{M}}
    \frac{|A\cap B|}{|A\cup B|}\in[0,1].
\end{equation*}

\mlayer{Distribution layer (reference-based, alignment-free).}
This layer compares label \emph{distributions} without any shot alignment,
sidestepping shot-count mismatch. For each field, per-episode label
histograms $P$ (AI) and $Q$ (director) are built over the closed label space
with $\varepsilon{=}10^{-6}$ smoothing and normalized (multi-label movement is
expanded per label). With $M=\tfrac12(P+Q)$ and base-2 logarithms,
\begin{equation*}
  \textsc{js}=\tfrac12 D_{\mathrm{KL}}(P\Vert M)
             +\tfrac12 D_{\mathrm{KL}}(Q\Vert M)\in[0,1],
\end{equation*}
computed as \textsc{sz-js}, \textsc{an-js}, \textsc{mv-js}. Lower means the
AI camera-language distribution is closer to the professional director's.

\mlayer{Reference-free layer.}
These metrics need no human reference and are computed on a single storyboard.

\paragraph{Entity out-degree (\textsc{eod}).}
Characters and props are CSV columns, so no NER is needed. The shot--entity
bipartite graph is projected onto a shot graph; for every pair $i<j$ the edge
weight is the number of shared entities. With $n$ shots,
\begin{equation*}
  \textsc{eod}=\frac{1}{n}\sum_{i<j}\bigl|\mathrm{ent}(i)\cap
    \mathrm{ent}(j)\bigr|,
\end{equation*}
measuring cross-shot entity carry-over~\cite{entitygraph2013}.

\paragraph{Adjacent cosine (\textsc{acos-m}, \textsc{acos-n}).}
With $\mathbf{v}_i$ the encoded description of shot $i$, we report the mean
and min of adjacent similarities $\mathbf{v}_i\!\cdot\!\mathbf{v}_{i+1}$,
measuring local coherence~\cite{cmlbench2025}.

\paragraph{NLI contradiction (\textsc{nli-cr}, \textsc{nli-mp}).}
Using the multilingual NLI model
\texttt{mDeBERTa-v3-base-xnli\dots-2mil7}~\cite{mdeberta_xnli2022}, each shot
$h_i$ is scored against its previous $k{=}3$ shots as premises, and the
shot-level contradiction score is the maximum:
\begin{equation}
  c_i=\max_{\,i-k\le j<i}\;
    P_{\mathrm{NLI}}(\text{contradiction}\mid h_i,h_j).
  \label{eq:nli}
\end{equation}
We then report \textsc{nli-mp}, the episode mean of $c_i$, and \textsc{nli-cr},
the fraction of shots with $c_i\ge 0.5$ (SummaC-style~\cite{summac2022}).

\begin{figure}[t]
  \centering
  \includegraphics[width=\columnwidth]{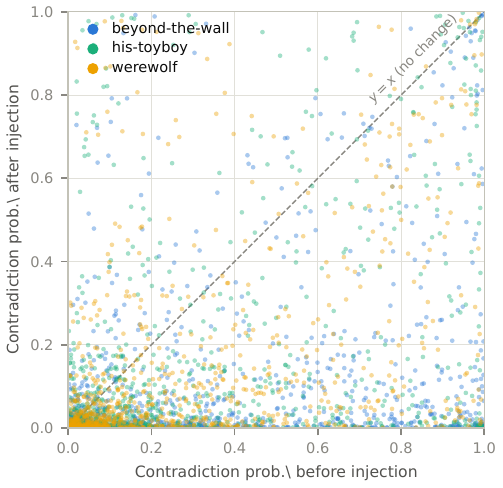}
  \Description{Scatter plot of before-injection versus after-injection
    contradiction probability for every repair-flagged shot, with most points
    falling below the diagonal and a dense band along the horizontal axis.}
  \caption{Targeted state-conditioned contradiction probability before
    vs.\ after memory injection for all 2{,}946 repair-flagged shots
    (three dramas $\times$ six backbones, one point per shot). Points
    below the $y{=}x$ diagonal are repaired; the mass collapsing onto the
    $x$-axis corresponds to the two-thirds contradiction drop of
    Table~\ref{tab:targeted}, and the sparse above-diagonal points are
    the residual failures discussed in
    Section~\ref{sec:exp_analysis}.}
  \label{fig:targeted_scatter}
\end{figure}

\mlayer{Targeted state-conditioned contradiction probe.}
The targeted probe measures repair only at the sparse sites the selective
filter flags. For each flagged modification we take the graph-expected entity
state as the NLI premise and score the shot description as hypothesis, before
and after injection. Over the flagged set we report
\begin{align*}
  \textsc{tc-p} &= \text{mean of }
    P_{\mathrm{NLI}}(\text{contradiction}\mid \text{expected state}, d),\\
  \textsc{tc-r} &= \text{fraction of flagged shots with } P_{\mathrm{NLI}}\ge 0.5,
\end{align*}
for $d\in\{\text{before},\text{after}\}$. Repair is effective when the
\emph{after} value drops well below the \emph{before} value.
Figure~\ref{fig:targeted_scatter} plots all 2{,}946 shot-level before/after
pairs behind Table~\ref{tab:targeted}.

\begin{figure}[t]
  \centering
  \includegraphics[width=\columnwidth]{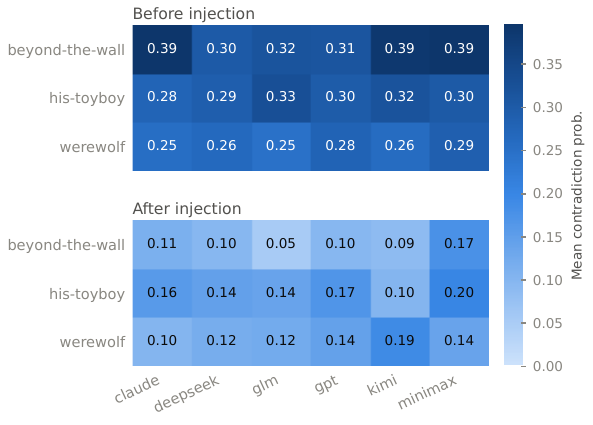}
  \Description{Two heatmaps of mean contradiction probability over three dramas
    by six backbones, before and after memory injection.}
  \caption{Mean targeted contradiction probability per drama--backbone cell,
    before (top) and after (bottom) memory injection. All 18 cells improve,
    from a $0.25$--$0.39$ band down to $0.05$--$0.20$
    (Section~\ref{sec:exp_analysis}).}
  \label{fig:targeted_heatmap}
\end{figure}

\begin{figure}[t]
  \centering
  \includegraphics[width=\columnwidth]{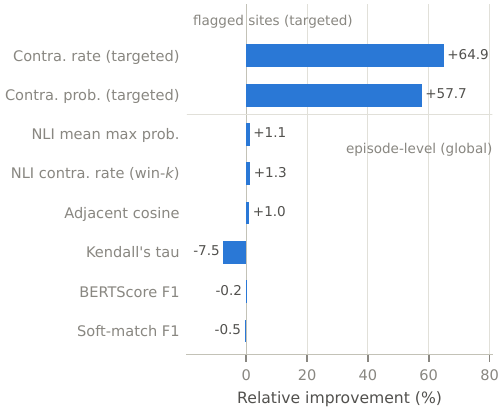}
  \Description{Horizontal bar chart contrasting large relative improvements on
    the two targeted metrics against near-zero shifts on six episode-level
    metrics.}
  \caption{Relative improvement at flagged sites versus at the episode level.
    The two targeted metrics gain $57.7\%$ and $64.9\%$; every global metric
    moves by at most $1.3\%$, the dilution effect of
    Section~\ref{sec:exp_analysis}.}
  \label{fig:dilution}
\end{figure}

\begin{figure}[t]
  \centering
  \includegraphics[width=\columnwidth]{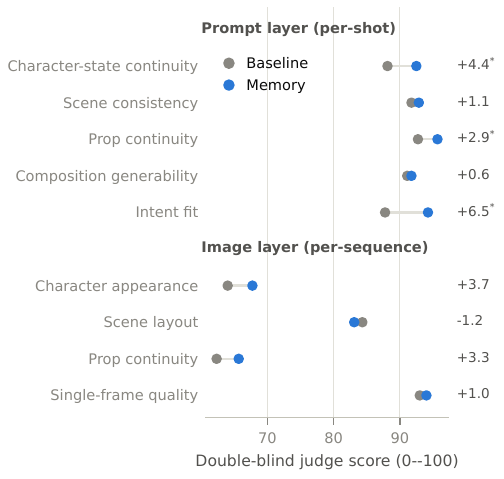}
  \Description{Dumbbell chart with one row per judged dimension, each row
    joining the baseline score to the memory score; prompt-layer rows show
    wide gaps and image-layer rows show narrow ones.}
  \caption{Double-blind judge scores by dimension (data of
    Table~\ref{tab:blind}): baseline (gray) vs.\ memory (blue), pooled
    over six backbones; $^{*}$ Holm-corrected $p<0.05$. The prompt layer
    shows large, significant gains on the dimensions injection edits,
    while image-layer deltas point the same way but are attenuated by the
    stochastic text-to-image stage (Q3).}
  \label{fig:blind_dumbbell}
\end{figure}

\mlayer{Image layer.}
Keyframes are scored with open vision checkpoints on CPU, with a sha1 feature
cache. Face crops use MediaPipe BlazeFace (confidence $0.5$, box expanded
$30\%$) with a full-image fallback when no face is found (detection rate
$0.48$). Metrics are computed only on shots that both conditions generated
successfully, so safety-filter refusals never bias the pairing.

\paragraph{DINOv2 similarities (\textsc{cref}, \textsc{sref}, \textsc{sself}).}
With DINOv2 (\texttt{facebook/\allowbreak dinov2-\allowbreak base}) CLS
embeddings, $L_2$-normalized:
\textsc{cref} is the mean cosine between a generated face crop and the
character reference faces; \textsc{sref} the mean full-image cosine between a
generated frame and its scene reference; \textsc{sself} the mean full-image
cosine between generated frames sharing a scene. All in $[-1,1]$, higher is
better.

\paragraph{Character self-consistency (\textsc{cself}).}
For frames sharing a character, we average the DreamSim
distance~\cite{dreamsim2023} (\texttt{dino\_vitb16}, distance $=1-\cos$).
Lower means a more stable cross-shot appearance.

\paragraph{CLIP-T (\textsc{clip-t}).}
Mean cosine between the CLIP (\texttt{clip-\allowbreak vit-\allowbreak
base-\allowbreak patch32}) image embedding and the text embedding of the
visual-description part of the keyframe prompt,
measuring image--prompt fidelity.

\paragraph{Aesthetic (\textsc{aes}).}
The LAION improved-aesthetic linear head~\cite{laion_aesthetic2022} on a CLIP
ViT-L/14 embedding, yielding a $1$--$10$ single-frame quality score.

\mlayer{Statistics.}
All baseline-vs-memory contrasts are paired at the episode level. We use the
Wilcoxon signed-rank test~\cite{wilcoxon1945} on the differences
$\mathrm{diff}=\text{memory}-\text{baseline}$ (fewer than six nonzero pairs
yields no $p$-value); the matched-pairs rank-biserial effect size is
$r=(W_{+}-W_{-})/(W_{+}+W_{-})$; and $p$-values are Holm-corrected within each
metric family at $\alpha{=}0.05$~\cite{holm1979} (\,$^{*}$ marks corrected
$p<0.05$).

\section{Online Deployment Architecture}
\label{app:deployment}

\begin{figure}[t]
  \centering
  \includegraphics[width=\columnwidth]{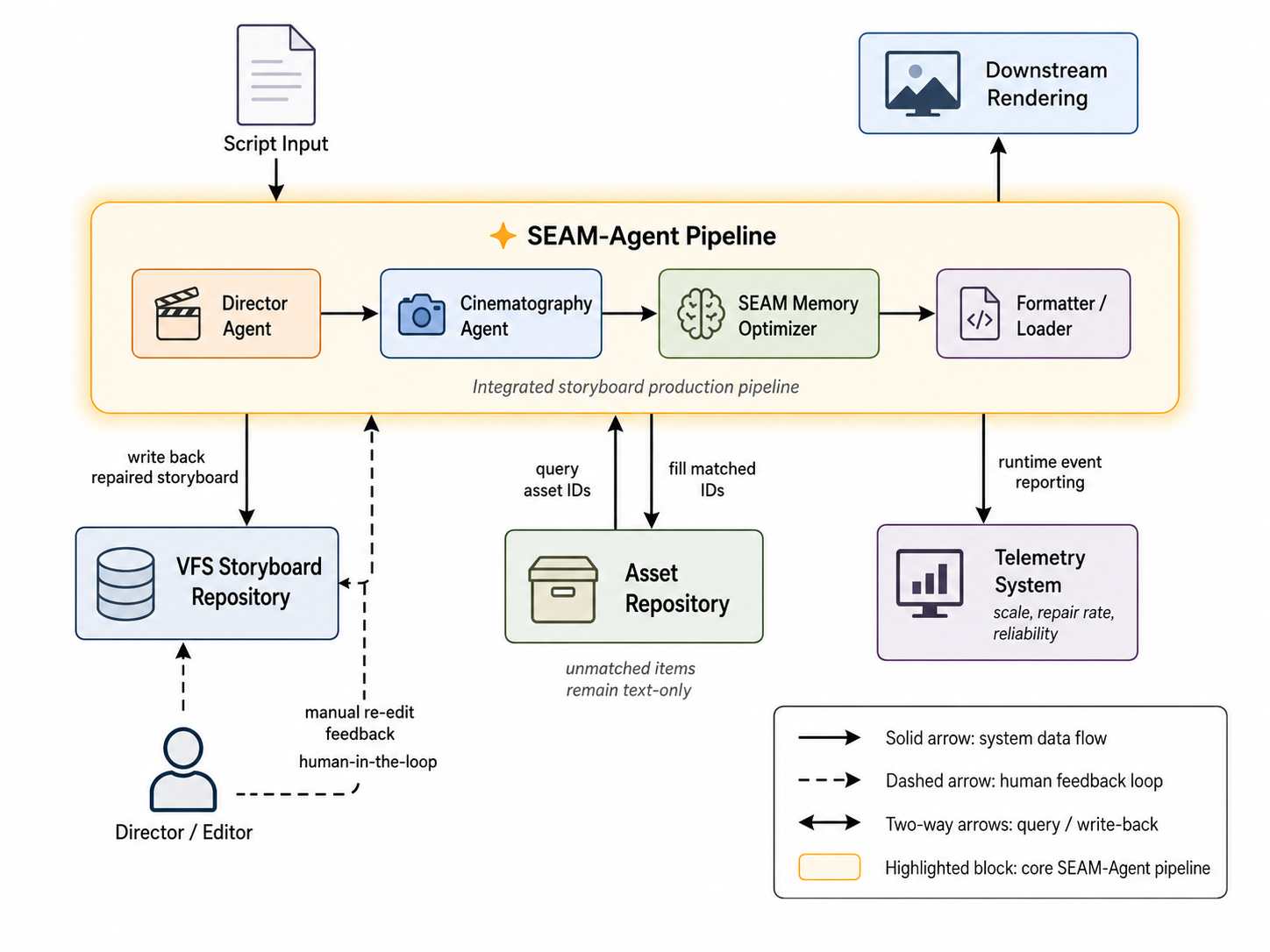}
  \Description{Architecture diagram of the production system: a script flows
    through the Director, Cinematographer, and SEAM Memory Optimizer agents to
    the Formatter/Loader and downstream rendering, with side connections to a
    shared asset repository, a storyboard repository that carries director
    feedback back into the pipeline, and a telemetry system.}
  \caption{Online deployment of \sysname as the memory-optimization stage of
    the \pipename storyboarding pipeline in CreativeFitting's production system.
    A script flows through the Director, Cinematographer, and SEAM Memory
    Optimizer agents to the Formatter/Loader and on to downstream rendering.
    Solid arrows are system data flow; the bidirectional arrows are asset query and
    write-back against the shared repository; dashed arrows are the
    feedback from directors and editors that keeps a human in the loop; the highlighted block
    is the core \pipename pipeline.}
  \label{fig:deployment}
\end{figure}

This appendix details the production pipeline behind the online run of
Section~\ref{sec:exp_online}. In CreativeFitting's system, \sysname is not a
standalone tool but the memory-optimization stage of \pipename, a multi-agent
storyboarding pipeline that turns a raw episode script into a
rendering-ready shotlist. Figure~\ref{fig:deployment} shows the full data flow.

A script enters the pipeline and passes through three agents in sequence. The
\emph{Director Agent} decomposes the script into shots and fixes pacing and
narrative intent; the \emph{Cinematographer Agent} assigns shot size, camera
angle, and movement, and resolves each shot's entities against the shared
\emph{Asset Repository}: it queries asset IDs and fills back the matched ones,
while unmatched items remain text-only and are never forced onto a wrong asset.
The \emph{SEAM Memory Optimizer} is where \sysname runs as a mandatory stage:
it extracts the multi-dimensional shot state, retrieves causally prior context
from the memory graph, filters it selectively, and rewrites the affected visual
descriptions, so continuity is repaired before the shotlist ever reaches a
renderer. Finally the \emph{Formatter/Loader} serializes the enriched shotlist
for \emph{Downstream Rendering}.

Two loops close the system around the directors who own the creative intent.
The repaired storyboard is written back to the \emph{VFS Storyboard Repository},
from which a director or editor may issue \emph{manual re-edit feedback}; this
path keeps a human in the loop, letting them override any stage without leaving the
pipeline. Separately, the Formatter/Loader reports runtime events to a
\emph{Telemetry System} that tracks production scale, repair rate, and
reliability. The acceptance figures of Section~\ref{sec:exp_online} are read
directly from the annotated shotlists rather than from this channel, so they
reflect the human decisions themselves and not aggregated monitoring counters
(Table~\ref{tab:online_adoption}).

\end{document}